\documentclass[letterpaper, 10 pt, journal, twoside]{IEEEtran}

\usepackage{amsmath}
\usepackage{mathtools}
\usepackage[linesnumbered,ruled,vlined]{algorithm2e}
\usepackage{amsfonts}
\usepackage{xr-hyper}
\usepackage{hyperref}
\usepackage{amssymb}
\usepackage{graphics} % for pdf, bitmapped graphics files
\usepackage{epsfig} % for postscript graphics files
\usepackage{mathptmx} % assumes new font selection scheme installed
\usepackage{times} % assumes new font selection scheme installed
\usepackage[linesnumbered,ruled,vlined]{algorithm2e}
\usepackage{subcaption}
\usepackage[utf8]{inputenc}
\usepackage{graphicx}
\usepackage{bm}
\usepackage{xcolor}
\usepackage{enumitem}
\usepackage{cleveref}
\usepackage{acronym}
\usepackage{tabularx}
\setlist{nolistsep}

\DeclareMathOperator*{\argmin}{arg\,min}

\acrodef{remp}[REMP]{reachability-expressive motion planning}
\acrodef{tamp}[TAMP]{Task and Motion Planning}

\makeatletter
\newcommand*{\addFileDependency}[1]{% argument=file name and extension
  \typeout{(#1)}
  \@addtofilelist{#1}
  \IfFileExists{#1}{}{\typeout{No file #1.}}
}
\makeatother

\begin{document}

\markboth{IEEE Robotics and Automation Letters. Preprint Version. Accepted January, 2022}
{Gao \MakeLowercase{\textit{et al.}}: Capability Calibration on Reachable Workspace for Human-Robot Collaboration} 

% Make room for more info lines in the \author command 
\author{Xiaofeng Gao$^{1}$, Luyao Yuan$^{1}$, Tianmin Shu$^{2}$, Hongjing Lu$^{3}$, Song-Chun Zhu$^{4}$%
\thanks{Manuscript received: September 9, 2021; Revised December 26, 2021; Accepted January 12, 2022.}%Use only for final RAL version
\thanks{This paper was recommended for publication by Editor Gentiane Venture upon evaluation of the Associate Editor and Reviewers' comments. This work was supported by DARPA XAI N66001-17-2-4029.} %Use only for final RAL version
\thanks{$^{1}$Xiaofeng Gao and Luyao Yuan are with UCLA Center for Vision, Cognition, Learning, and Autonomy (VCLA), Los Angeles, CA 90095, USA
        {\tt\footnotesize (e-mail: xfgao@ucla.edu, yuanluyao@ucla.edu})}%
\thanks{$^{2}$Tianmin Shu is with Department of Brain and Cognitive Sciences, Massachusetts Institute of Technology, Cambridge, MA 02139, USA
        {\tt\footnotesize (e-mail: tshu@mit.edu)}}%
\thanks{$^{3}$Hongjing Lu is with UCLA Department of Psychology, Los Angeles, CA 90095, USA
        {\tt\footnotesize (e-mail: hongjing@ucla.edu)}}%
\thanks{$^{4}$Song-Chun Zhu is with Beijing Institute for General Artificial Intelligence (BIGAI), Beijing 100080, China, with School of Artificial Intelligence and Institute for Artificial Intelligence, Peking University, Beijing 100871, China, and also with Department of Automation, Tsinghua University, Beijing 100871, China
        {\tt\footnotesize (e-mail: sczhu@bigai.ai)}}%
\thanks{Digital Object Identifier (DOI): see top of this page.}
}
% Use only for final RAL version.

\title{Show Me What You Can Do: Capability Calibration on Reachable Workspace for Human-Robot Collaboration}

\maketitle
% \thispagestyle{empty}
% \pagestyle{empty}

%%%%%%%%%%%%%%%%%%%%%%%%%%%%%%%%%%%%%%%%%%%%%%%%%%%%%%%%%%%%%%%%%%%%%%%%%%%%%%%%
\begin{abstract}
  Aligning humans' assessment of what a robot can do with its true capability is crucial for establishing a common ground between human and robot partners when they collaborate on a joint task. In this work, we propose an approach to calibrate humans' estimate of a robot's reachable workspace through a small number of demonstrations \textit{before} collaboration. We develop a novel motion planning method, \ac{remp}, which jointly optimizes the physical cost and the expressiveness of robot motion to reveal the robot's reachability to a human observer. Our experiments with human participants demonstrate that a short calibration using \ac{remp} can effectively align a non-expert user’s estimation of the robot’s reachability with its true capacity. We show that this calibration procedure not only results in better user perception, but also promotes more efficient human-robot collaborations in a subsequent joint task. \footnote{Code and demos are available at \url{https://xfgao.github.io/calib2022ral/}.}

\end{abstract}


\begin{figure*}
    \centering
    \begin{subfigure}{0.18\textwidth}
      \centering
      \includegraphics[width=\textwidth]{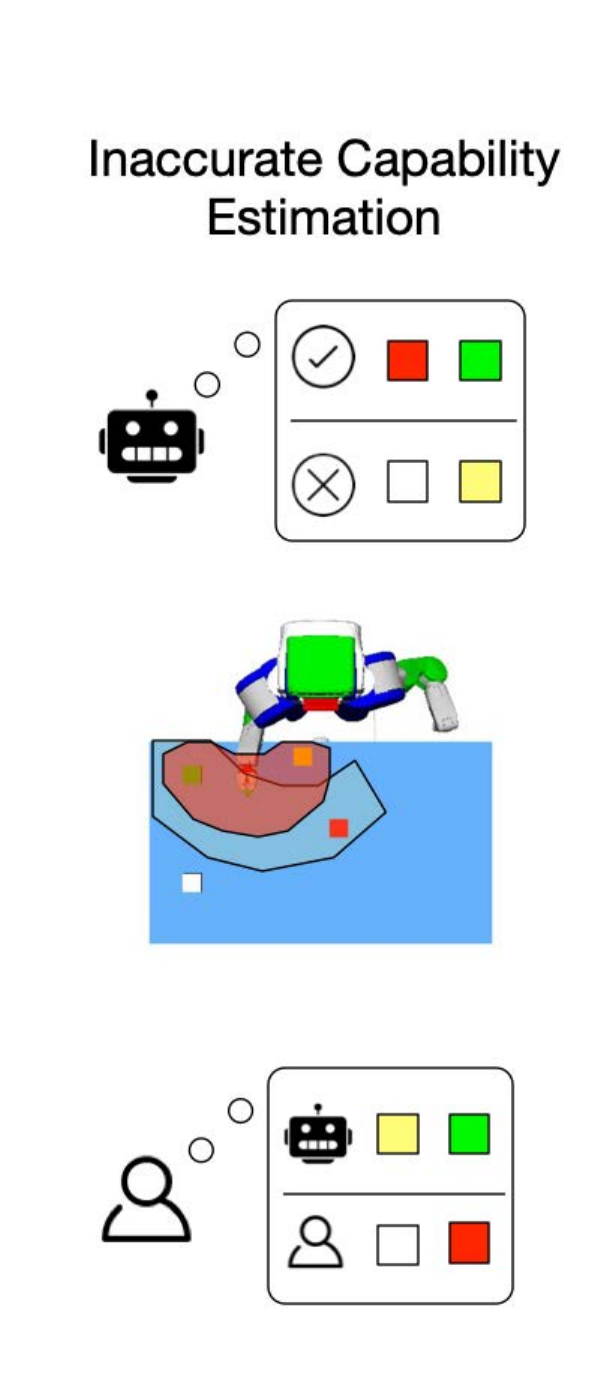}
      \vspace{-10pt}
      \caption{Inaccurate capability estimation can lead to failure in collaboration.}
    \end{subfigure}
    \quad
    \begin{subfigure}{0.78\textwidth}
      \centering
      \includegraphics[width=\textwidth]{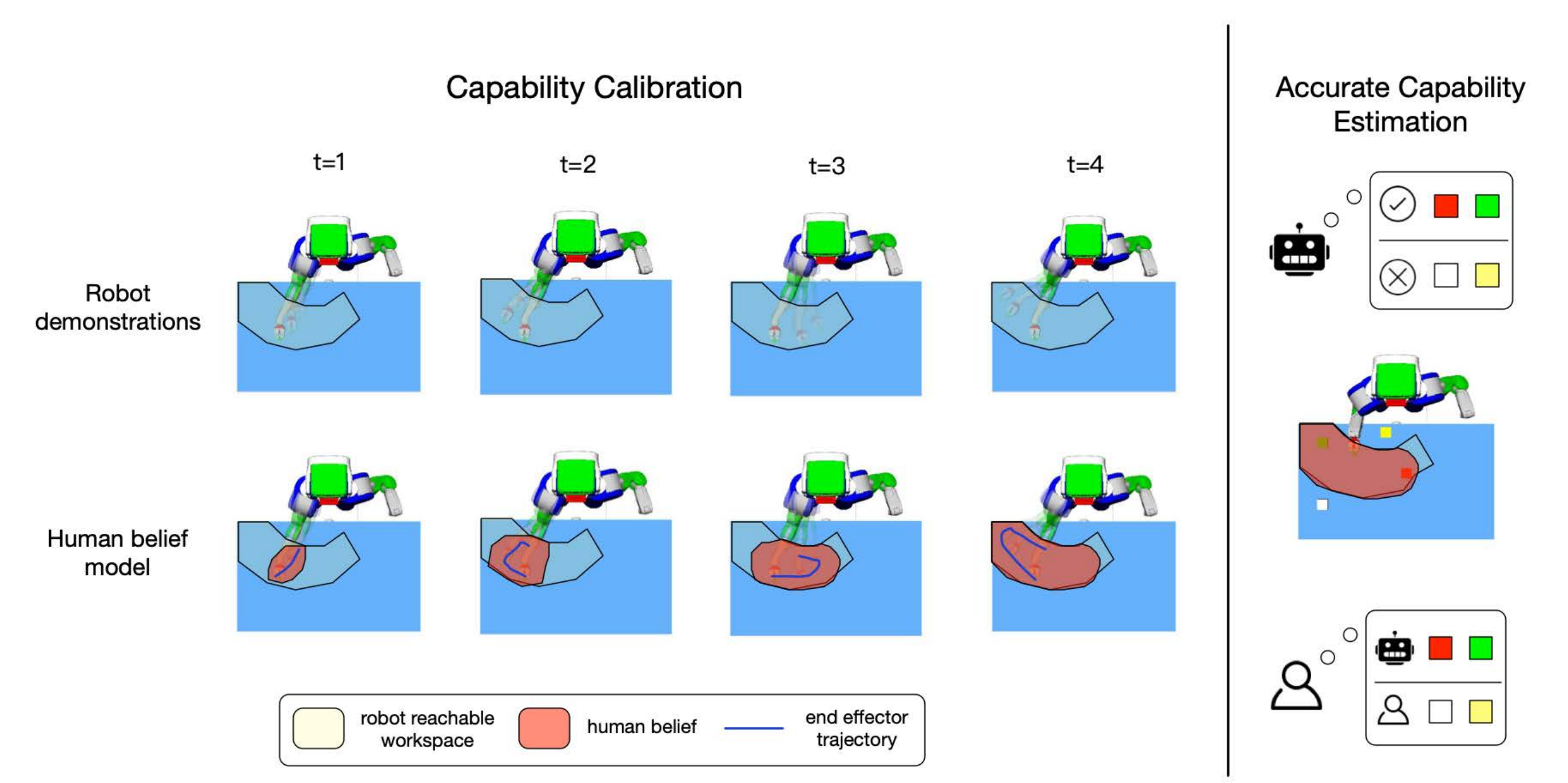}
      \vspace{-10pt}
      \caption{In this example, the human is supposed to pick up the white and the yellow cubes and let the robot collect the red and the green ones.}
    \end{subfigure}
    \vspace{-2pt}
    \caption{(a) Consider a collaborative table clearing task, where the robot has a limited capability and cannot reach the yellow and white objects. Inaccurate estimation of the robot's reachable workspace would harm collaboration: users who incorrectly estimate that the robot can reach the yellow object would assign it to the robot, resulting in a worse teaming performance. (b) We propose capability calibration, where the robot uses its motion to demonstrate its capability before collaboration. }
    \label{fig:intro}
    \vspace{-15pt}
\end{figure*}

\section{Introduction}\label{sec:intro}
\IEEEPARstart{O}{ne} of the main challenges in Human-Robot Interaction is that the capacity of the robot perceived by the human partner may not be consistent with its actual capacity \cite{stoffregen1999perceiving, powers2006advisor, fussell2008people}. Such discrepancy may lead to overuse or misuse of the robot. Particularly, in an ad-hoc teaming setting where humans do not have prior experience with their robot partners, the consequence caused by such discrepancy could be detrimental to the team collaboration~\cite{Albrecht2017Reasoning}.

In this work, our key insights to address this challenge are two-fold: i) humans' perception of the capability of a robot can be calibrated by observing its behavior, e.g., robot demonstrating its motion trajectories in pursuit of certain goals, and ii) calibrating the perceived robot capability improves the quality of subsequent human-robot collaboration.

We focus on a case study as shown in \Cref{fig:intro}, where a human user and a robot share the same workspace, and they must take turns picking up all objects as fast as possible. As the robot can only reach part of the workspace due to its mechanical limits, the human partner needs to pick up the objects that the robot can not reach to achieve maximum efficiency in completing this joint task. We introduce capability calibration as shown in \Cref{fig:intro}b, where we allow the robot to show a small number of demonstrations. After watching each demonstration, the human can estimate the robot's capability accordingly. The goal is to come up with motion plans to pragmatically demonstrate the robot's capability.

 To achieve a sample-efficient calibration, we propose reachability-expressive motion planning (\ac{remp}), a novel planning algorithm that models perceived robot capability as a human's belief over a robot's reachable workspace, and integrates the belief update into motion planning by introducing an additional cost in trajectory optimization. As a result, \ac{remp} can generate a series of expressive trajectories for different robots to showcase their reachability to users. We conducted a user study in which participants i) first observed several robot demonstrations, then ii) estimated where the robot could reach, and iii) proceeded to work with the same robot in a joint task: picking up all objects in the shared workspace as fast as possible. We find that i) \ac{remp} significantly increases the accuracy of humans' reachability estimation, ii) the subsequent human-robot collaboration benefits from a successful calibration, iii) users perceive the robot as more predictable and reliable.  
 \vspace{-3pt}

% %===============================================================================

\section{Related Work}
% \textbf{Perceived robot capability.} There have been various works studying how humans perceive a robot's capability. \cite{cha2015perceived} investigated the effects of robot speed and speech on perceived capability. \cite{nikolaidis2017game} used a game-theoretic approach to model human's expectations of the robot capability over time. As perceived robot capability plays an important part in trust in automation~\cite{muir1994trust}, previous works have also studied how capability inference affects human's trust and reliance on the robot \cite{chen2018planning, huang2018establishing, xie2019robot}. More recently, \cite{lee2020getting} formulated the intent and capability calibration problem as a Bayes Adaptive POMDP, and assumed that the agents model each other's ability based on experience counts of action outcomes. Unlike the previous work which focused on capability models on discrete action space, this work, to the best of our knowledge, is the first to integrate perceived capability models into motion planning. %By generating expressive trajectories, the robot can improve human's understanding of its kinematic constraints and avoid ineffective collaboration. 

\noindent\textbf{Perceived robot capability.} Various works have studied how humans perceive a robot's capability, differentiating between social and physical capabilities \cite{cha2015perceived}. In prior work, social capabilities were defined as a robot’s ability to communicate and interact with humans \cite{jung2013engaging}, and physical capabilities were defined on a set of tasks a robot can successfully perform \cite{robinson2021smooth}, such as lifting different objects on the table \cite{nikolaidis2017game, chen2018planning}, searching and firefighting in various weather and fire conditions \cite{xie2019robot}. In these works, robots' capabilities of different tasks are estimated separately based on experience counts of action outcomes -- a higher success rate indicates a stronger capability in a task \cite{lee2020getting}. Since the capability modeled in these works are highly task dependent, the user's knowledge of a robot's capability in one task can not be easily generalized to the knowledge of its capability in other tasks. In contrast, our work focuses on physical capabilities that serve as a basis for a robot to achieve success in a wide range of tasks. In particular, we focus on \textit{reachability}, which is one of the most fundamental physical capabilities for robots. By understanding a robot's reachability, users can better assess its overall capability in various tasks where reaching is involved. Given an arbitrary task, the user can decide whether the robot can successfully perform it based on perceived reachability. 

\noindent\textbf{Robot expressive motions.} When deploying robots in real-world settings that are beyond factory environments, functional motions only designed to accomplish tasks are inadequate for human users to correctly understand the robots and establish effective collaborations~\cite{venture2019robot}. It is equally important to convey the rationality and intent of a robot through its motion \cite{szafir2014communication, lemasurier2021methods}. To generate such motions, prior work formulated and optimized the legibility of trajectories via functional gradient descent~\cite{dragan2013generating,stulp2015facilitating}. Similar ideas were also adopted to study robots' expression of emotion~\cite{felis2015optimal} and style~\cite{liu2005learning}. To express robot (in)capability, prior work used repetitive motions, either generated by simple heuristics \cite{nicolescu2001learning} or hand-crafted for each task \cite{takayama2011expressing}. In contrast, \cite{kwon2018expressing} proposed a trajectory optimization-based method that maximizes the similarity between motions and would-be successful executions. Our work takes one further step in this direction: we i) model how humans update their beliefs of the capability of a robot given the observed robot motions and ii) integrate the belief update process into trajectory optimization to generate new motions that can optimally improve humans' beliefs. 

\vspace{-5pt}
% %===============================================================================

\section{Capability Calibration}\label{sec:calibration}
We propose a capability calibration framework (as shown in \Cref{fig:intro}b) to align a human user's understanding of a robot's capability with the ground truth, where the user can watch a small number of demonstrations of her robot partner before they work together. In this section, we introduce our approach to generate such demonstrations that can efficiently reveal the robot's reachability. We show how this calibration could be applied to collaboration in Section~\ref{sec:collaboration}. 

 \begin{figure*}
    \centering
    \includegraphics[width=\textwidth]{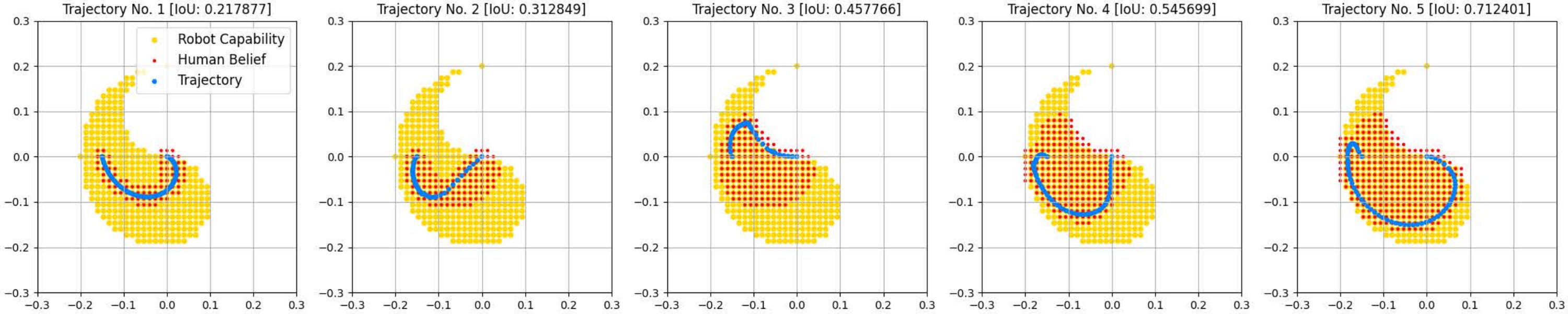}
    \caption{Simulated human estimation of robot A's reachability map, after observing each demonstration generated by \Cref{algo:remp_k}, measured by Intersection of Union (IoU) between the human estimation and the ground truth. Robot A is a 2-link arm with link lengths 0.1. }
    \vspace{-15pt}
    \label{fig:sim_belief_map}
\end{figure*}

\vspace{-3pt}
\subsection{Calibrating Reachable Workspace}
In this sub-section, we define the reachability calibration task. In \Cref{sec:belief}, we describe how human belief would be updated before a new trajectory can be generated. In \cref{sec:planning}, we propose \ac{remp}, which enables the robot to generate one trajectory showing its reachable workspace based on a simulated human belief that models what the human has already known about the robot.  In \Cref{sec:TAMP}, we reify our capability calibration framework by combining \ac{remp} and task planning. We begin with some notations.

The robot's ground truth reachability is defined as  $f: \mathcal{X}_{ws} \rightarrow \{0, 1\} $, i.e. whether a target position $x$ in the workspace $\mathcal{X}_{ws}$ is reachable by the end-effector according to the robot's kinematic constraints. Meanwhile, we assume the human is maintaining a belief $b_{h}^{t}: \mathcal{X}_{ws} \rightarrow [0, 1]$, modeling how likely a target is reachable after observing a robot trajectory $\xi^{t}_{1:N} \in \Xi$ with length $N$ at time $t \in \{1...T\}$. After observing all the robot demonstrations, human's final belief would become $b_h^{T}=\tau (b^{0}_{h}, Z)$, with $\tau$ denoting the belief transition from the initial guess $b^{0}_{h}$ to the final $b_h^{T}$ with all seen demonstrations $Z=\xi^{1:T}$. In addition, we define $\mathcal{X}_{rs}\subseteq\mathcal{X}_{ws}$ as the robot's reachable workspace.  $\phi_{ee}: \mathcal{Q} \rightarrow \mathcal{X}_{rs} $ is the forward kinematic function of the end-effector, generating its position given a configuration.

Using notations above, we can formalize capability calibration as an optimizing problem over a set of trajectories, $Z$, whose cardinality may not necessarily be known in advance. The goal is to reduce the mismatch between the robot's ground truth capability and the user's final belief:
\begin{align}
    &\argmin_{Z\in\Xi^*}Cost(Z)\label{eq:calibration_optimization}\\
    s.t. \sum_{x\in\mathcal{X}_{ws}}\Big|&\tau\big(b_h^0(x), Z\big)-f(x)\Big| < \epsilon,\nonumber
    \vspace{-3pt}
\end{align}
where $\Xi^*$ uses the Kleene star to represent all possible sequences of robot motions, $Cost(\cdot):\Xi^*\rightarrow\mathbb{R}$ is a function to evaluate the overall cost for trajectories. The condition means the user has a reachability estimation close enough to the robot's true capability.

One intuitive cost is the total length of all the trajectories in $Z$, optimizing which is equivalent to minimizing the cardinality of $Z$ when the trajectories all have similar length. In this paper, we maintain the homogeneity of trajectories by further regulating the start configurations of all trajectories to be in a set of configurations $S \subset \mathcal{Q}_{0}$ and target positions in a set of positions $G \subset \mathcal{X}_{rs}$. 

% Mathematically, we have
% \begin{align}
%     Cost(Z) = \sum_{\xi\in Z}\frac{1}{\mathbf{1}\big(\phi_{ee}(\xi_{N})\in G\big)\mathbf{1}(\xi_{1} \in S)}.
% \end{align}
% Namely, $Cost(Z)=|Z|$ if all the trajectories satisfy the starting configuration and target position regulation, otherwise $\infty$.

Due to the size of motion space, an exact solution to \cref{eq:calibration_optimization} is intractable. Thus, we adopt an incremental update: we keep generating new trajectories $\xi^{t}$ until the user's belief is sufficiently aligned with the robot's capability. Every time we want to expand $Z$, we first select a pair of starting configuration and target position $(q^t, x_r)\in S \times G$ and then generate a motion using it. We term the former as the task planning problem and the latter as motion planning. 

% In next section, we define the human belief model. In \cref{sec:planning}, we propose \ac{remp} to plan motion \textbf{given} a starting configuration and a target position. In \cref{sec:TAMP}, we integrate \ac{remp} with task planning.
\vspace{-5pt}

\subsection{Human Belief Model} \label{sec:belief}
% \vspace{-2pt}
% \noindent \textbf{Human belief update model.} 
Our objective is to make people without any knowledge about robotics easily understand the true capacities of a robot. Thus, our human belief model attempts to capture what a novice user may think about a robot’s reachability after watching its trajectories. We assume human updates its belief on an interested point $x$ in the workspace after observing a new robot trajectory $\xi$. Intuitively, if a point is close to the visited positions in an observed trajectory, the human observer would consider it more likely to be reachable. We model the belief update process as an iterative Bayesian inference beginning from a uniform prior:
\begin{equation}
	\tau(b^t, \xi^t)(x) = b^{t+1}_{h}(x) \propto b^{t}_{h}(x) p(\xi^t|x)
\label{eq:belief}
\end{equation}
and $p(\xi^t|x)=e^{-\gamma d(\phi(\xi^t), x)}$, where $d(\phi(\xi), x)$ captures the distance between the trajectory $\xi$ and the interested position $x$. The hyperparameter $\gamma$ defines how much the human extrapolates the observed trajectory to the points nearby: a large $\gamma$ means that such extrapolation mainly happens to the point which is very close to the trajectory. In particular, we use the end-effector position $\phi_{ee}$ as the feature, and compute the squared euclidean distance between the interested position and the closest end-effector position in the trajectory: 
\begin{equation}
\vspace{-3pt}
    d(\phi(\xi), x) = \min_{i} ||\phi_{ee}(\xi_{i})-x||^{2}.
    \label{eq:dist_func}
\vspace{-2pt}
\end{equation}

The design of our distance function is motivated by the fact that, given a trajectory, it is straightforward for users to focus on the robot's end-effector which is central to the task, while trying to estimate its reachable workspace. 

\subsection{REMP: Reachability-Expressive Motion Planning} \label{sec:planning}
% \vspace{-2pt}

\begin{algorithm}[t]
	\caption{\ac{remp}}
	\label{algo:REMP}
	Given a target position $x_{r}$ and a starting configuration $q^{t}$, human belief $b_t$\;
	Generate trajectory $\xi^{t}$ based on $b^{t}_{h}$, $q^t$, \Cref{eq:motion} \;
	Update human belief $b^{t+1}_{h}$ using $\xi^{t}$, \Cref{eq:belief} \;
	\Return $b_h^{t+1}$, $\xi^{t}$
% 	\vspace{-2pt}
\end{algorithm}

Expressing robot reachability is more than randomly moving the end-effector to somewhere in its reachable workspace. Our insight is that it is essential to understand what the human already knows or does not know about the robot, so that every demonstration can communicate as much information to the human as possible. We believe this can be formulated as an optimization problem: finding a new trajectory that would minimize the misalignment between the ground truth reachability and human's updated estimation. We capture the misalignment using a cost function $c(\xi, b_{h}^{t}, f)$ and formulate the optimization problem as the following:
\vspace{-10pt}

\begin{equation}
\label{eq:motion}
\begin{aligned} 
    \xi^t = \argmin_{\xi} \quad & c(\xi, b_{h}^{t}, f) + \frac{1}{\lambda} \sum_{i=1}^{N} || \xi_{i+1} - \xi_{i} ||^{2}, \\
     \textrm{subject to} \quad & \phi_{ee}(\xi_{n}) = x_{r}, \textrm{collision-free} (\xi).
\end{aligned}
\end{equation}

The first term is an expressiveness cost and the second term is a smoothness cost commonly seen in trajectory optimization. The trajectory at the $t$-th step is generated by minimizing the sum of the two costs, subjecting to a constraint that requires the end effector to reach a target position $x_{r}$ at the end of the trajectory.

Assuming each point in the trajectory contributes to the cost independently, we design the cost function based on a value $v_{i}(\xi_{i}, b_{h}^{t}, f)$, which represents the degree of alignment between human's estimation and the robot's ground truth reachable workspace:
\vspace{-10pt}

\begin{equation}
\vspace{-3pt}
\begin{aligned} \label{eq:cost_b}
	c_{b}(\xi, b_{h}^{t}, f) &= \alpha \sum_{i=1}^{N} {v_{i}(\xi_{i}, b_{h}^{t}, f)} \\
	&= \alpha \sum_{i=1}^{N} e^{\beta \big(b_{h}^{t}(\phi_{ee}(\xi_{i}))-f(\phi_{ee}(\xi_{i})) \big)} \\
\end{aligned}
\vspace{-3pt}
\end{equation}

A small value $v_{i}$ suggests that the human observer is underestimating the robot's capability at $\xi_{i}$. In that case, we want to facilitate calibration by encouraging the robot to move to $\xi_{i}$. On the other hand, we would see a large $v_{i}$ if the human is over-estimating the capability. In that case, it is beneficial for the robot to avoid reaching points near $\xi_{i}$. The hyperparameter $\alpha$ and $\beta$ control how aggressive the trajectory would be in expressing the capability. We call this cost function \bm{$c_{b}$}, which captures human updated belief. Note that the intuition is if the observer previously underestimates the reachability of a point $x$, $b^t_h(x) - f(x)$ will be negative and give low cost for trajectories covering $x$. Hence, trajectories passing through underestimated points are more likely to be chosen. Trajectories including overestimated points, on the contrary, have larger costs and are less likely to be selected.

\noindent\textbf{Static human model.}  Our key intuition is the human would update its belief of the robot's reachability after observing each trajectory. To test it, we also design a fixed cost function as baseline, assuming an underlying uniform belief model $ \forall x, \ b_{static}(x) = b_{0}. \label{eq:static} $
The corresponding cost function under the assumption of a static human model is \bm{$c_{s}$}. Note that this baseline generates functional motions that solely aim to finish the physical task of reaching the target. We envision that in reality, users may also learn from these physical motions the robot's capability by interacting with the robot on some tasks, but such learning is not as efficient as the learning in a dedicated calibration phase. 

\begin{figure*}[t]
\begin{subfigure}{\textwidth}
    \centering
	\includegraphics[width=\textwidth]{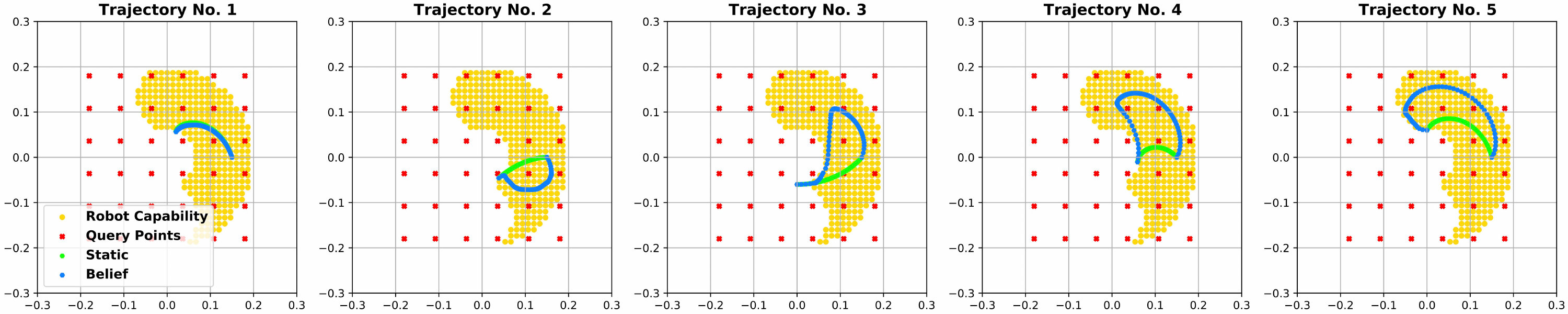}
	\caption{Robot B is a 2-link arm with link lengths $0.13$ and $0.07$.}
	\label{fig:2link_trajs}
	\vspace{5pt}
\end{subfigure}

\begin{subfigure}{\textwidth}
    \centering
	\includegraphics[width=\textwidth]{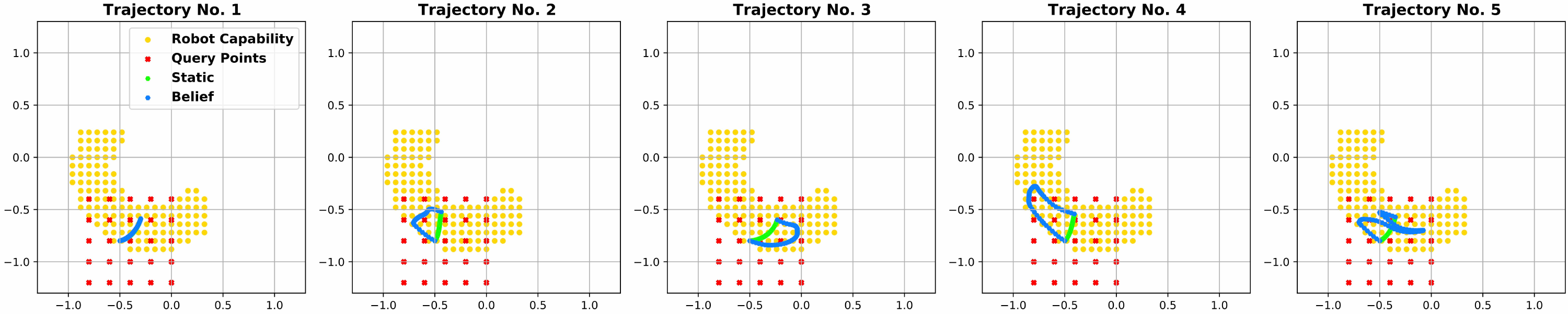}
	\caption{Robot C is a PR2 robot. In this work, we consider the reachable workspace of its right arm.}
	\label{fig:pr2_trajs}
\end{subfigure}
\vspace{-3pt}
\caption{Visualization of the robot reachable workspace and the trajectories generated by cost function $c_b$ (\textit{belief}) and $c_s$ (\textit{static}). (a) and (b) show the results for Robot B and Robot C respectively. It can be seen that the \textit{belief} trajectories cover broader regions of the reachable workspace and new trajectories tend to visit areas that haven't been covered by their predecessors. The red dots, corresponding to \Cref{fig:eval}, represent the points we use to query the users in our experiments.}
\vspace{-15pt}
\end{figure*}

\vspace{-2pt}
\subsection{Generating Reachability-Expressive Trajectories}\label{sec:implementation}
% \vspace{-2pt}

\textbf{Implementation.} We implemented our framework using TrajOpt \cite{schulman2013finding} on two kinds of simulated robots in OpenRAVE \cite{10.5555/2125842}, including a manipulator with 2 links and a PR2 robot. For the 2-link arm, we manipulated its joint limits and link lengths to allow it to have a variety of two-dimensional reachable workspaces. These serve as testing cases for our framework, as we want to study how well the framework copes with reachable workspaces of different sizes and shapes. For the PR2 robot, we didn't do such manipulations since the goal here is to see how practical it is to apply the framework to real robot manipulators. Without loss of generality, we focus on the right arm of the PR2 robot. In practice, we use grid search to find hyperparameters that generate trajectories to maximize the accuracy of reachability estimation in simulation, as described in \Cref{sec:sim-result}. 

\begin{algorithm}[t]
    \caption{\ac{remp}-T}
    \label{algo:remp_k}
    Given a list of target position and starting configuration pairs $(x, q)^{1:T}$ and human belief $b_h$\;
    \For{$t=1, \hdots, T$}
    {
        $b_h, \xi^t$ = \ac{remp}($x^t, q^t, b_h$)
    }
    \Return $b_h, \xi^{1:T}$
\end{algorithm}

\begin{algorithm}[t]
	\caption{Calibration on Reachable Workspace}
	\label{algo:task-planning-full}
	Given a set of target positions $G$, starting configurations $S$, initial human belief $b_0$\;
	\For{$t = 1, 2, \hdots, $}
	{
    	$\forall x, \ b_{h}^{0}(x) \leftarrow b_{0}$\;
    	Let $\sigma(\zeta) = \sum_{x\in\mathcal{X}_{ws}}\big|f(x) -$ \ac{remp}-$T(\zeta, b_h^0)[b_h]\big|$\;
    	$\zeta^*_t \leftarrow \argmin_{\zeta\in (G \times S)^t} \sigma(\zeta)$ \;\label{alg:TAMP_full_inner_argmin} 
    	$\sigma_t \leftarrow \sigma(\zeta^*_t)$\;
    	\If{$\sigma_{t}-\sigma_{t-1}<\epsilon$}
    	{\Return \ac{remp}-$T(\zeta^*_t, b_h^0)$}
	}
\end{algorithm}

\begin{algorithm}[t]
	\caption{Calibration with Fixed $T$}
	\label{algo:task-planning}
	Given a set of target positions $G =\{x_1, \hdots, x_N\}$, number of trajectories $T$, starting configuration set $S$, initial human belief $b_0$\;
	$\forall x, \ b_{h}^{0}(x) \leftarrow b_{0}$, $\delta \leftarrow \infty$\;
	\For{$\kappa\in T-\text{combination}(G\times S)$}
	{
	    $b_h\leftarrow b_h^0$\;
	    \For{$t\gets$ 1 \KwTo $T$}
	    {
	        $b_h, \hat{\xi}^t \leftarrow$ \ac{remp}($\kappa^t(G), \kappa^t(S), b_h$)\;\label{algo:task-planning:start}
	    }
	    $\sigma = \sum_{x\in \mathcal{X}_{ws}}|b_h(x) - f(x)|$\;
	    \If{$\sigma < \delta$}
	    {
	        $\delta \leftarrow \sigma$\;
	        $\xi^{1:T} \leftarrow \hat{\xi}^{1:T}$\;
	    }\label{algo:task-planning:dist}
	}
	\Return $\xi^{1:T}$
% 	\vspace{-3pt}
\end{algorithm}

\noindent\textbf{Qualitative behaviors.} 
\Cref{fig:sim_belief_map}, \Cref{fig:2link_trajs} and \Cref{fig:pr2_trajs} show the trajectories generated by the cost functions $c_{b}$ and $c_{s}$ for the robots by running \ac{remp} iteratively using the updated belief, following \Cref{algo:remp_k}. As both $c_{b}$ and $c_{s}$ assume a uniform belief on the robot's reachable workspace at the beginning, the first trajectories generated by these cost functions are almost identical. Starting from the second trajectories, we find that the ones generated using $c_{b}$ can cover a large part of the robot's reachable workspace. On the contrary, trajectories generated by $c_{s}$ are more sensitive to the physical cost. Overall, It is clear that \ac{remp} accommodates human belief at each time step and tries to traverse uncovered regions to better express the robot's reachability. 
\vspace{-3pt}

\subsection{Planning for Start and Target Pairs}\label{sec:TAMP}
\vspace{-1pt}
We have shown how \ac{remp} can generate an expressive trajectory \textit{given} a starting configuration and a target position. For a better capability calibration, we also want to optimize the number of trajectories as well as the sequence of starting configurations and target positions. As outlined in \cref{algo:task-planning-full}, this could be achieved by \ac{tamp} \cite{kaelbling2011hierarchical}, where the plans of start and target pairs come from task planning and the trajectories for a given pair comes from \ac{remp}. In \cref{algo:task-planning-full}, we solve \eqref{eq:calibration_optimization} in an incremental manner. Namely, we keep increasing the cardinality of $Z$ until user's belief is aligned with the robot's actual capability. For each size of $Z$, we find the best sequence of starting configurations from $S$ and target positions from $G$ (\cref{alg:TAMP_full_inner_argmin} of \cref{algo:task-planning-full}). To avoid trajectories that are too short or uninformative, we set $S$ as the set of configurations  near the neutral configuration of the robot and $G$ as the set of positions far away from the neutral end effector positions:

\vspace{-15pt}
\begin{align}
    S = \{ q, \forall q \ | q - q_{neutral} | < a_{1} \}
\end{align}
\vspace{-20pt}
\begin{align}
    G = \{ x, \forall x \ \min_{q \in S} | x - \phi_{ee}(q) | > a_{2} \}
\end{align}
\vspace{-10pt}

\begin{figure*}[t]
    \begin{subfigure}{0.30\textwidth}
        \centering
        \includegraphics[width = \textwidth]{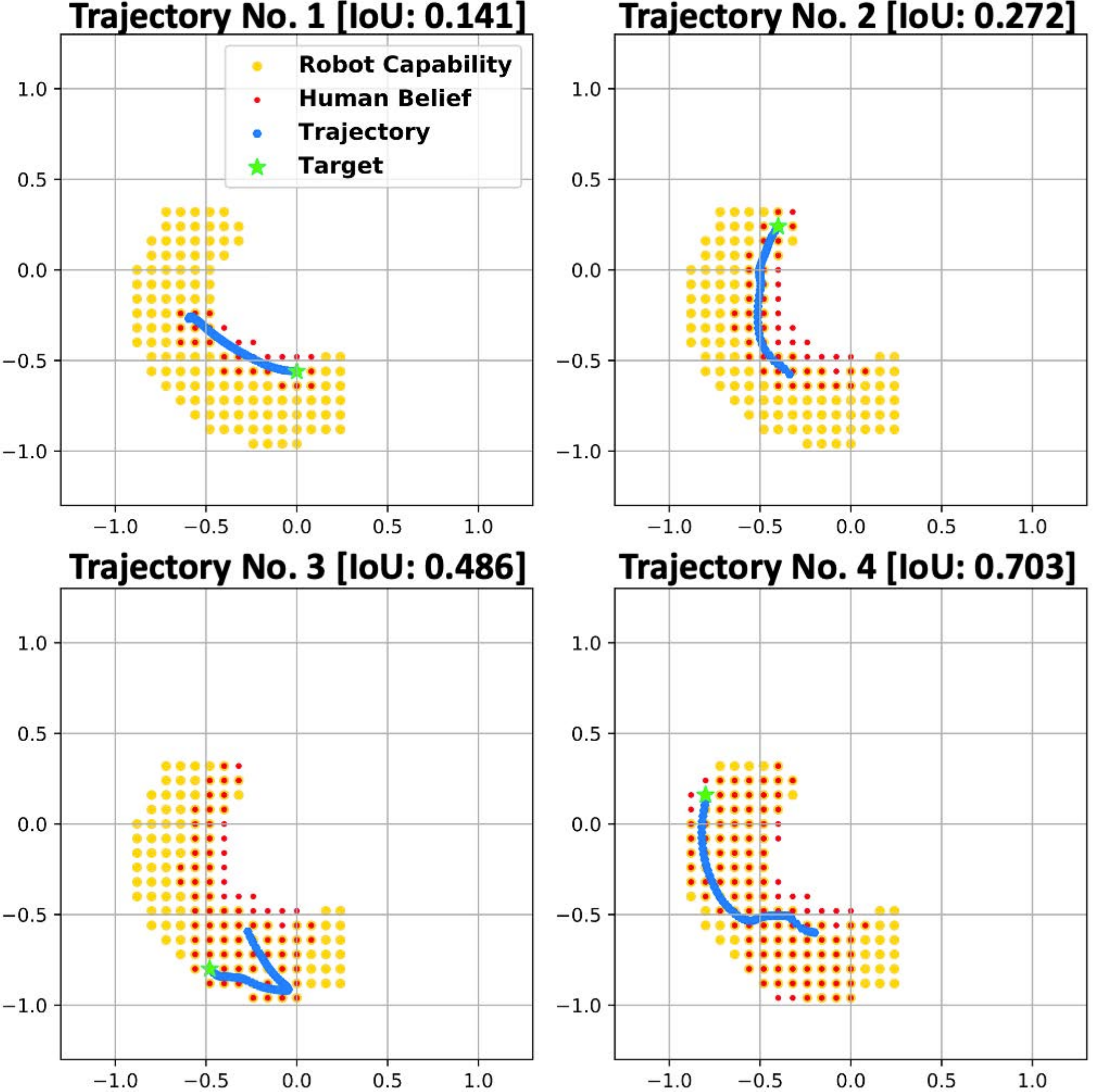}
        \caption{Trajectories generated by task and motion planning and the simulated reachability estimation given observed trajectories.}
        \label{fig:multiple-targets}
        \vspace{-5pt}
    \end{subfigure}
    \begin{subfigure}{0.69\textwidth}
        \begin{subfigure}{\textwidth}
            \centering
            \vspace{-5pt}
            \includegraphics[width=0.8\textwidth]{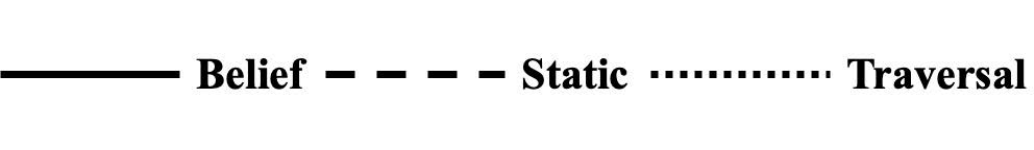}
            \vspace{-15pt}
        \end{subfigure}
        
        %\par\medskip
        \begin{subfigure}{0.5\textwidth}
        \renewcommand\thesubfigure{\alph{subfigure}1}
          \centering
          \includegraphics[trim=0 0 0 0, clip,width=\textwidth]{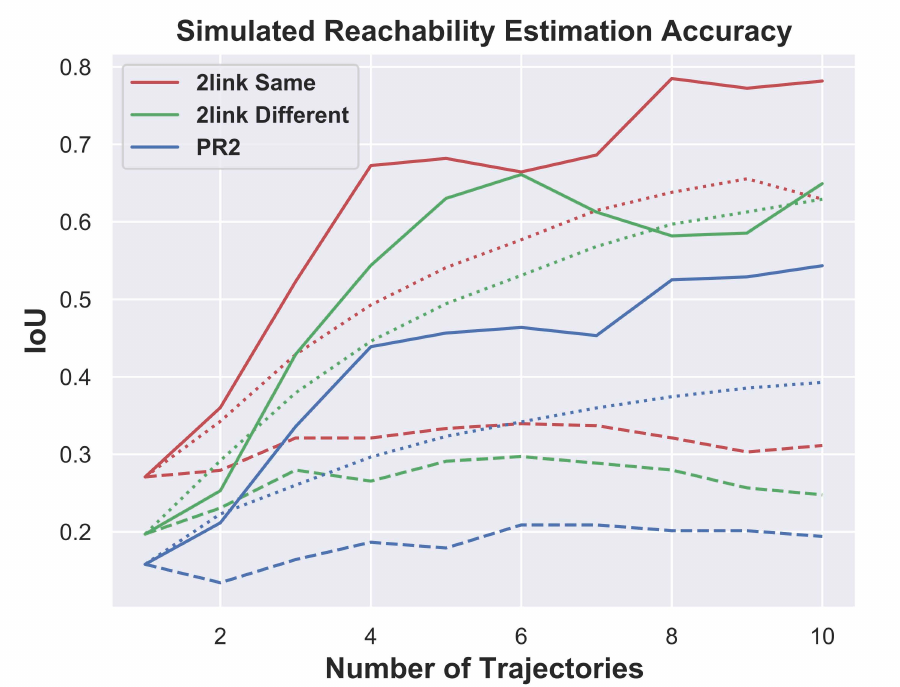}
          \vspace{-5pt}
          \caption{Simulated reachability estimation accuracy, measured by Intersection of Union between the human reachability estimation and the ground truth. Higher value indicates better estimation.}
          \label{fig:sim_iou}
        \end{subfigure}%
        ~
        % \par\medskip\medskip
        \begin{subfigure}{0.5\textwidth}
        \addtocounter{subfigure}{-1}
      \renewcommand\thesubfigure{\alph{subfigure}2}
            \centering
          \includegraphics[trim=0 0 0 0, clip, width=\textwidth]{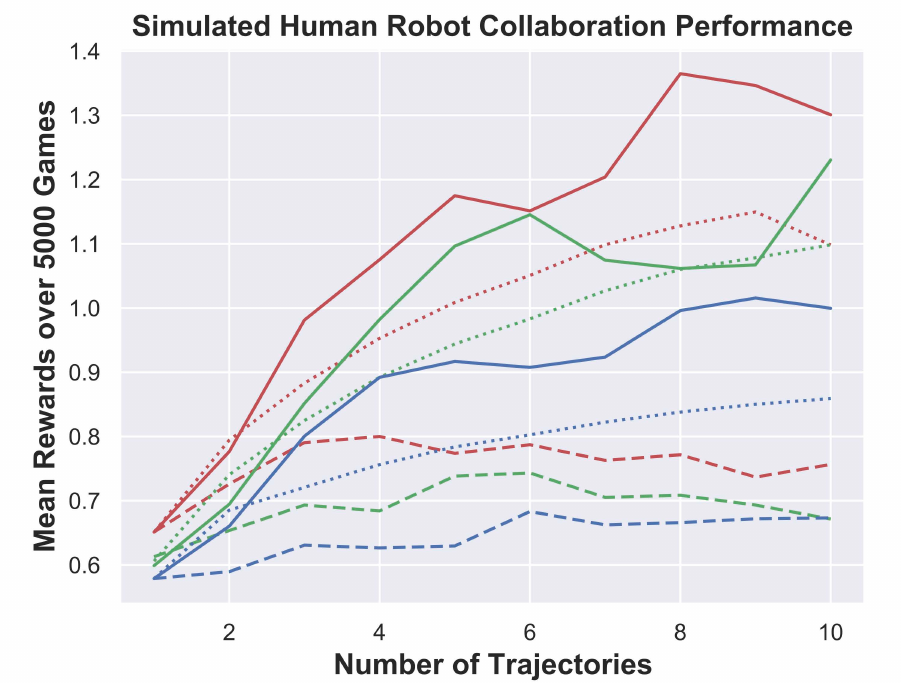}
          \vspace{-2pt}
          \caption{Simulated human-robot collaboration performance measured by averaged rewards acquired by the group. Every point on the curves for \textit{traversal} is the mean of 100 trajectories.}
          \label{fig:sim_reward}
        %   \vspace{-5pt}
        \end{subfigure}%
        \addtocounter{subfigure}{-1}
        \vspace{-2pt}
    \end{subfigure}
    \caption{\textbf{(a)} Combining \ac{remp} with task planning, we can optimize the starting and target positions for better calibration. \textbf{(b)} Simulation results of reachability estimation and collaboration performance. Starting and target positions are chosen greedily.}
    \label{fig:sim_coop_performance}
    \vspace{-10pt}
\end{figure*}

Finding $Z$ incrementally, despite giving the exact optimum, can be time consuming. In practice, rather than demonstrating to humans constantly until converge, we can pre-define $T$ to a reasonable number and find the optimal set of trajectories. In \cref{algo:task-planning}, we assume fixed number of trajectories. We start from a uniform prior for the human belief, and update the belief w.r.t. Eq.~\eqref{eq:belief}. \Cref{fig:multiple-targets} depicts an example to optimize 4 trajectories that start from different configurations and reach different targets. Note that \cref{algo:task-planning} plans by enumerating all possible combinations, but any stochastic planning approaches can be used to further accommodate resource constraints and task scalability. 
% %===============================================================================

\section{Applying \ac{remp} to Human-Robot Collaboration}\label{sec:collaboration}
In this section, we discuss how to apply \ac{remp} to human-robot collaboration \textit{after} the calibration. 
\vspace{-3pt}

\subsection{Collaborative Table Clearing} \label{sec:task}
We design a human-robot collaboration task in a table clearing scenario, where some objects are scattered on a table and a robot can assist the human with the object collection. The human and the robot take turns picking up the objects. In each step, the human collects first and the robot collects one of the remaining objects. The human can reach all of the objects, while the robot can only reach a subset of them. To finish the task as quickly as possible, the human and the robot need to split the work wisely, so that, in each round, the robot has some objects to pick up. The reward is calculated by the number of objects picked up and the time penalty.
\vspace{-3pt}

\subsection{Human and Robot Policy} 
After observing robot expressive demonstrations and updating the belief with Eq.~\eqref{eq:belief}, the human is assumed to act in an approximately rational way with respect to the current estimation of the robot capability, $b^t_h$. We use a Boltzmann noisily-rational human decision model~\cite{morgenstern1953theory}, assuming the human is more likely to help the robot with its unreachable objects based on the human's current reachability estimation. Since we want to emphasize the effect of the calibration,  we  use  a  simple  uniform  robot  policy  in  the simulation,  i.e.,  it  would  randomly  pick  up  objects  it  can reach, and do nothing if no objects are reachable. %Details of the human and robot policy can be found in \Cref{sup:sec:MDP} of the supplementary material.
\vspace{-1pt}

\subsection{Simulation Results} \label{sec:sim-result}
Using the behavior model described in the previous sections, we simulated with 3 robots A, B and C with different configurations and reachability: (i) A is a 2-link arm where each link is of equal length, (ii) B is a 2-link arm where the length of its first link (0.13) is larger than the length of the second (0.07), (iii) C is a PR2 robot. The \textit{belief} and \textit{static} methods in the legend correspond to the definition in \Cref{sec:belief}. In addition, we implemented a \textit{traversal} baseline, where the robot moves its end-effector to traverse the workspace to demonstrate its reachability. From the starting position, the end effector moves to unreached waypoints in its reachable workspace one by one. The number of waypoints we sampled corresponds to the number of trajectories in \textit{belief} and \textit{static}.

\Cref{fig:sim_iou} and \Cref{fig:sim_reward} shows the quantitative results of capability calibration and human-robot collaboration. The result suggests that as the robot shows more demonstrations, the human has a better understanding of its capability and collaborates with it more effectively for all baselines. Looking at the sample efficiency, we notice that without modeling human belief changes, the improvement is quite slow and limited: a large number of demonstrations need to be observed before calibration is achieved. On the contrary, trajectories generated by our proposed \ac{remp} algorithm keep providing new information to the user. As a result, the user's estimation accuracy increases much faster for \textit{belief} compared to the baselines. There is fluctuation when many trajectories are shown, due to the limited memory of our human model.
\vspace{-3pt}

%Details of the game setup can be found in supplementary \Cref{sup:sec:MDP}. 

% %===============================================================================

\begin{figure}[t]
	\centering
	\begin{subfigure}{0.21\textwidth}
      \centering
      \includegraphics[width=\textwidth]{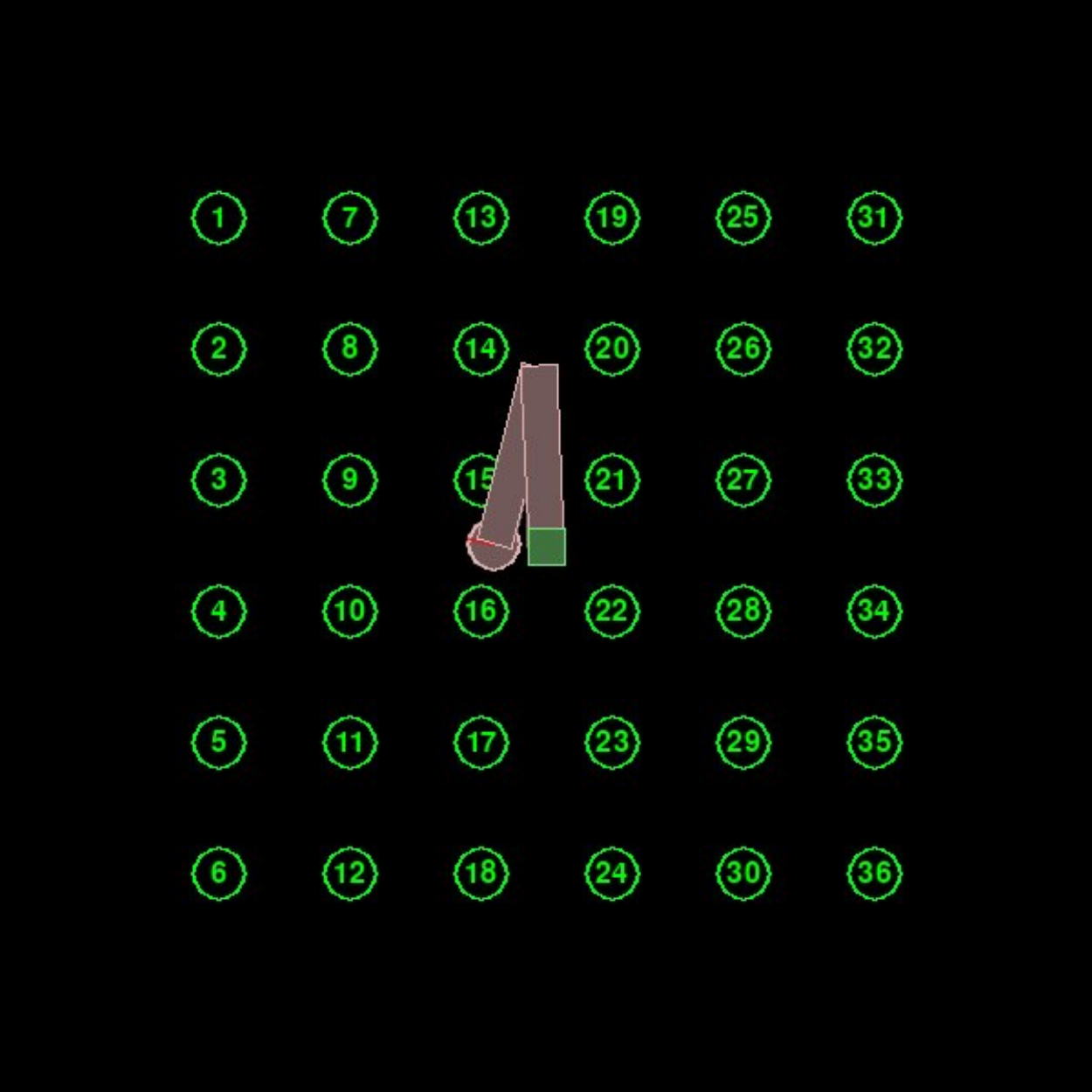}
    \caption{36 points for 2-link arms.}
    \end{subfigure}
    ~
    \quad
	\begin{subfigure}{0.21\textwidth}
      \centering
      \includegraphics[width=\textwidth]{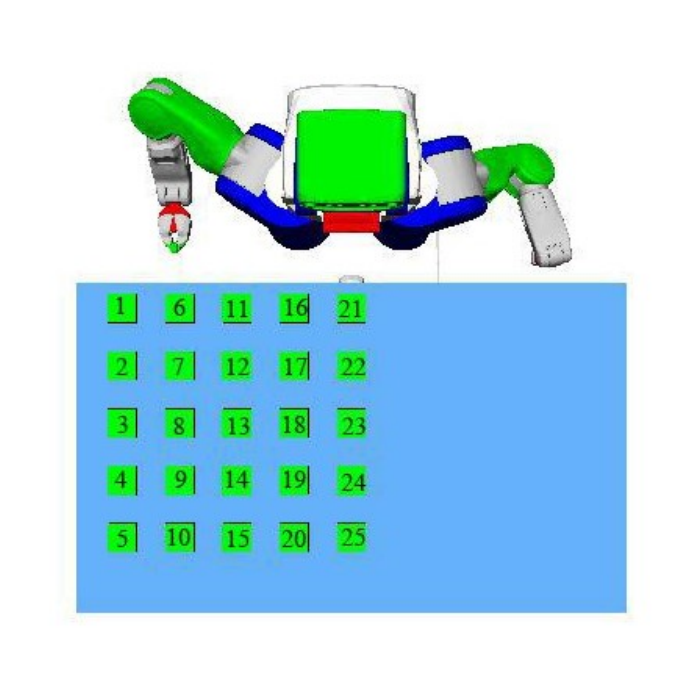}
    \caption{25 points for PR2.}
    \end{subfigure}
    \quad
	\caption{To evaluate users' estimation of the robot's reachable workspace, we sample query points in the workspace and ask users to select points that they think the robot's end effector can reach. These points correspond to the red dots in \Cref{fig:2link_trajs} and \Cref{fig:pr2_trajs}.}
	\label{fig:eval}
\end{figure}

\section{User Study}
As we have shown the effectiveness of \ac{remp} in simulation, we now turn to investigate how much it helps users work with robots in a user study. This study was certified as exempt from IRB review per 45 CFR 46.104 category 3 by the UCLA Institutional Review Board on 9/4/2020.
\vspace{-5pt}
\subsection{Experiment Design}\label{sec:exp-design}

\noindent{\textbf{Participants.}}
We recruited 202 participants (37\% Female, median age 34) from Amazon Mechanical Turk. 

\noindent{{\textbf{Materials.}}
During the study, participants interact with the three robots in the simulation as described in \Cref{sec:sim-result}}. We measure how well participants can understand the robot's capability and how such understanding can help them in the collaboration, as well as their self-reported perception of the robot. To measure \textbf{capability understanding}, we ask users to choose positions that they think the robot can reach from a number of object queries, as shown in \Cref{fig:eval}. We record their selections and compare them with the ground truth. For \textbf{collaboration task performance}, we use the accumulated reward of the team as a measure. To measure the \textbf{perception of the robot}, we ask participants to rate statements listed in  \Cref{tab:survey} on a 7-point Likert scale labeled from "strongly agree" to "strongly disagree", after they have finished interaction with a robot. Inspired by \cite{madsen2000measuring}, the statements shown in \Cref{tab:survey} are designed to evaluate their subjective understanding of the robot in different aspects, including reachability, predictability, reliance and trust. 

\begin{table}[t]
\caption{Survey statements to evaluate reachability, predictability, reliability and trust toward robots.}
\label{tab:survey}
\begin{tabularx}{0.49\textwidth}{l} 
 \hline
1. It is easy to tell where the robot’s hand can reach. \\
2. The robot behaves in a predictable manner.\\
3. I can rely on the robot to function properly despite its limited capability.\\
4. I trust the robot.\\
  \hline
\end{tabularx}
\vspace{3pt}
\end{table}

\begin{figure*}
    \begin{subfigure}[t]{0.32\textwidth}
      \centering
      \includegraphics[width=\textwidth]{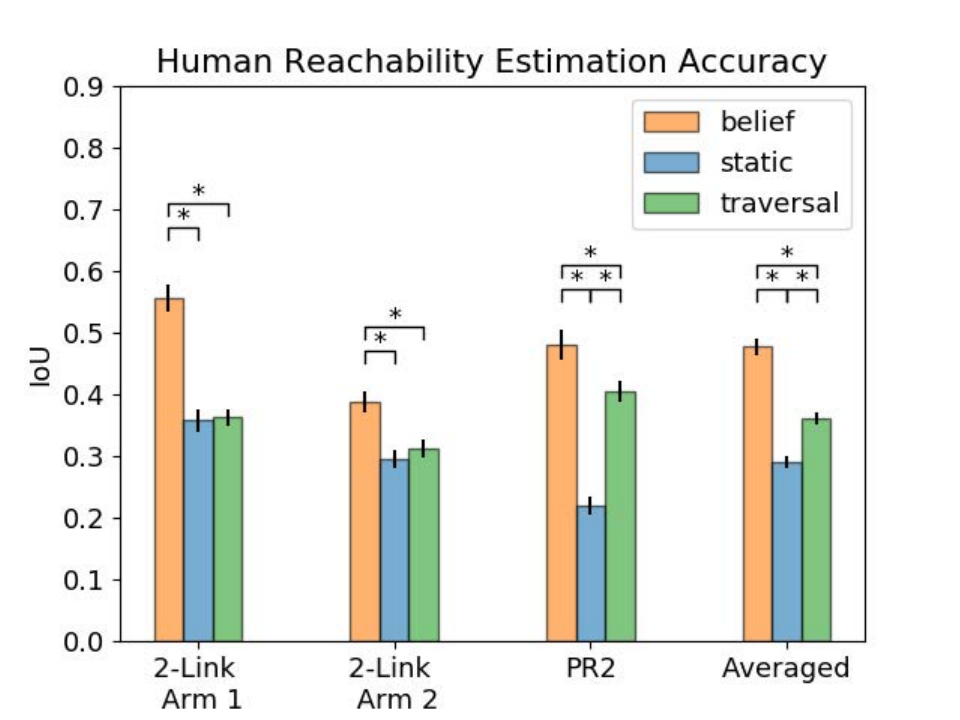}
    \caption{Intersection of Union between the human reachability estimation and the ground truth. A higher value indicates better estimation.}
      \label{fig:user_reachability}
    \end{subfigure}%
    \quad
    \begin{subfigure}[t]{0.32\textwidth}
        \centering
        \includegraphics[width=\textwidth]{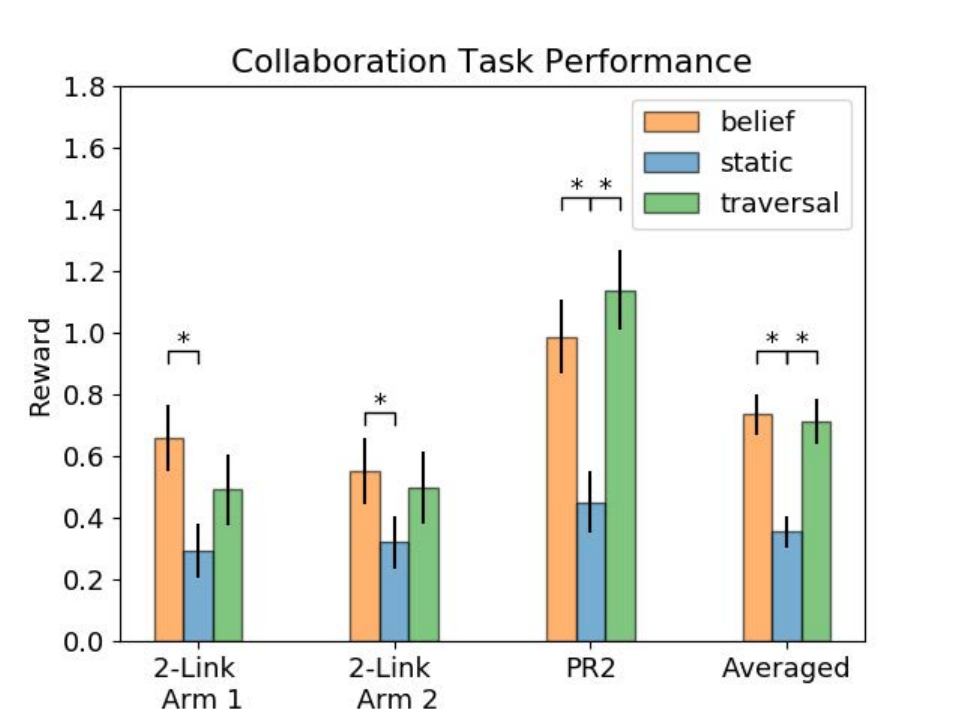}
        \caption{The human-robot team performance in the collaboration task. A higher value indicates a higher reward.}
        \label{fig:task_performance}
    \end{subfigure}%
    \quad
    \begin{subfigure}[t]{0.32\textwidth}
        \centering
      \includegraphics[width=\textwidth]{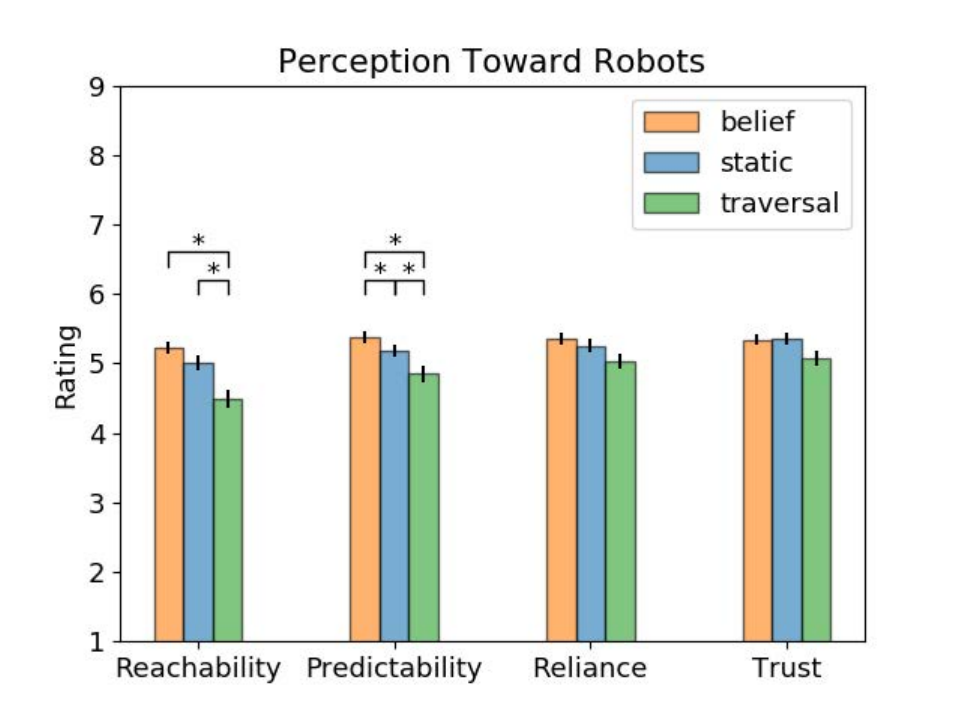}
        \caption{Users' ratings toward the Likert statements in \cref{tab:survey}. A higher rating indicates higher confidence.}
        \label{fig:survey}
    \end{subfigure}
    \vspace{-2pt}
    \caption{User study results. Here we report means and standard errors. * indicates statistical significant pairs ($p<.05$).}
    \vspace{-15pt}
\end{figure*}

\noindent\textbf{Calibration task.} 
In the calibration task, the participant would be randomly assigned to an experiment group, and observe $T$ demonstrations. Based on the simulations results in \Cref{fig:sim_iou}, we believe showing more demonstrations would generally lead to a better calibration. Considering the limited time of participants in our online study, however, we cannot use an arbitrarily large $T$. From simulation, we witness the most significant improvement during the first $4$ trajectories, thus we control the number of demonstrations $T=4$ in practice. After seeing all demonstrations, participants are asked to estimate the robot's reachable workspace by choosing positions that they think the robot can reach from a number of object queries, as shown in \Cref{fig:eval}. 

\noindent\textbf{Collaboration task.} 
In the collaboration task, participants are asked to perform an online table clearing task together with the same robot they have just been calibrated. As discussed in \Cref{sec:task}, the task required the team to clear all four objects on the table. During each time step, the participant would pick up an object first before the robot makes its decision. The team would get rewarded based on how fast they take all the objects. Since two of the objects cannot be reached by the robot, to get the maximum accumulated reward (+2), the participant needs to pick up objects that cannot be reached by the robot. Failure to do so would result in the team getting a lower reward (0). 

\noindent\textbf{Experiment conditions.} 
Like simulations, we varied types of motion users observed in the user study, i.e., the \textit{belief}, \textit{static} and \textit{traversal} methods defined in Section~\ref{sec:sim-result}.

\noindent\textbf{Design.} 
The robot types are within-subject: participants interacted with all three robots. Demonstrations are between-subject: participants only saw demonstrations from one of the three experiment conditions when interacting with a robot. 

\noindent\textbf{Procedure.}
 After a brief introduction, each participant is asked to interact with three robots A, B and C in random order. The purpose is to see how robot's physical configurations affect capability perception. During the interaction with each robot, the participants would first go through a calibration task before collaborating with the same robot on the table clearing task.

\noindent\textbf{Hypotheses.}
We hypothesized the capability calibration framework benefits the users in the following aspects:

\noindent\textbf{H1:} Participants going through capability calibration in the \textit{belief} condition would have a better understanding of the robot's capability, compared to those in the other conditions.

\noindent\textbf{H2:} Teams in the \textit{belief} condition would perform better in the collaboration tasks than those in the other conditions.

\noindent\textbf{H3:} Participants in the \textit{belief} condition would have a more positive perception of the robot, compared to those in the other conditions.

\vspace{-8pt}

\subsection{Result and Analysis}
\vspace{-3pt}
\noindent\textbf{Capability understanding.}
We first analyzed the accuracy of the user's estimation of the robot's reachable workspace, by computing the intersection of union (IoU) between their responses and the ground truth. We performed a Kruskal–Wallis H-test of the IoUs using the type of motion independent variables. As a result, we found a significant effect for the motion (${\chi}^{2}(2, 603)=113.52, p<.001$). A post-hoc analysis with Mann-Whitney U test revealed that all three conditions are different from each other, with \textit{belief} significantly better than \textit{static} ($p<.001$) and \textit{traversal} ($p<.001$). This confirms our hypothesis \textbf{H1}. \Cref{fig:user_reachability} shows the accuracy of participants' reachability estimation w.r.t different robots. On average, \textit{belief} performs 65\% better than \textit{static} and 32\% better than \textit{traversal}. Compared to the simulation results in \Cref{fig:sim_coop_performance}, the user study result follows relatively the same order for different conditions. 

\noindent\textbf{Task performance.}
We also analyzed the collaboration task performance. A Kruskal–Wallis H-test indicates that there is a statistically significant effect of the accumulated rewards between conditions (${\chi}^{2}(2, 603)=22.62, p<.001$). The post-hoc Mann-Whitney U test showed a significant difference between \textit{belief} and \textit{static} ($p<.001$). This partially supports our hypothesis \textbf{H2}. We didn't observe a significant difference between \textit{belief} and \textit{traversal}. \Cref{fig:task_performance} shows the task performance for different robots in three conditions. The Pearson correlation coefficient between reachability estimation and collaboration performance is $r=.203$ with p-value smaller than $0.001$, indicating a positive correlation. This validates that calibrating perceived robot capability benefits the collaboration performance. Surprisingly, users in the \textit{traversal} condition have a slightly higher reward when collaborating with the PR2 robot compared with those in \textit{static}, even if their reachability estimation is less accurate, although the difference is not significant. This is probably due to the stochastic nature of the \textit{traversal} baseline and specific object locations in our collaboration task.

\noindent\textbf{Perception of robots.}
Finally, we analyzed participants' perception toward robots. Running a Kruskal–Wallis H-test, we found significant effects for reachability (${\chi}^{2}(2, 603)=20.39, p<.001$) and predictability (${\chi}^{2}(2, 603)=11.30, p=.003$). The post-hoc Mann-Whitney U test revealed significant difference between \textit{belief} and \textit{traversal} for reachability ($p<.001$) and predictability ($p<.001$), confirming \textbf{H3}. Overall, users tended to prefer \textit{belief} over \textit{static}, and \textit{static} over \textit{traversal}. This is unexpected for reachability, considering the fact that users are actually better at predicting robots' reachability in \textit{traversal} than in \textit{static}. The Pearson correlation coefficient between reachability rating and prediction accuracy is $r=.109$, indicating a weak positive correlation. Similarly, we observe a weak correlation $r=.019$ between self-reported reliance and users' actual ability to rely on the robot during collaboration. This suggests that there may be a discrepancy between the users’ self-reported capability understanding and what they actually know about the robot.

In summary, we found that users in the \textit{belief} condition had the most accurate estimation of the robots' capability, and reported the robots in this condition as the most reliable, the most predictable, and the easiest to understand among all three conditions. Moreover, users working with the \textit{belief} robots achieved a higher reward than those working with the \textit{static} robots did. These objective and subjective results together suggest that our approach has an overall advantage for improving humans' understanding of robots as well as the quality of collaboration over the baselines.

\vspace{-3pt}
% %===============================================================================

\section{Conclusion}
\label{sec:discussion}
We have proposed an expressive robot motion planning algorithm, \ac{remp}, which can generate informative trajectories by integrate human belief update into trajectory optimization. Our experiments show that our approach can efficiently calibrate a user's perception of a robot's reachability and consequently improve human-robot collaboration.

In this work, we focused on the robot's spatial reachability. As reaching is one of the most basic tasks in human-robot interaction, we believe understanding reachability would greatly help users understand robot capacities in more complex tasks. Thus we view our work as a successful first step towards a more general capability calibration setting. In the future, it is possible to extend \ac{remp} for other capabilities.  Our current work treats predictability and reliability as separate measures from trust. Considering the multidimensional nature of trust, better instruments can be used for a more comprehensive trust measure \cite{malle2021multidimensional}. Also, due to online experiment constraints, we only investigated the reachability calibration problem on a 2D plane. We intend to generalize our algorithm to 3D environment in future work. 

%For future work, we also intend to explore online capability calibration during the collaboration process. Another promising direction is t in this o develop multi-modal demonstrations, including gestures, gazes and verbal communication \cite{breazeal2005effects}. We are also interested in using a variety of other modalities  (e.g. speech, gesture) \cite{breazeal2005effects} as ways of communicating capabilities.  

\iffalse
\section*{ACKNOWLEDGMENT}

The preferred spelling of the word ÒacknowledgmentÓ in America is without an ÒeÓ after the ÒgÓ. Avoid the stilted expression, ÒOne of us (R. B. G.) thanks . . .Ó  Instead, try ÒR. B. G. thanksÓ. Put sponsor acknowledgments in the unnumbered footnote on the first page.
\fi

%%%%%%%%%%%%%%%%%%%%%%%%%%%%%%%%%%%%%%%%%%%%%%%%%%%%%%%%%%%%%%%%%%%%%%%%%%%%%%%%
\bibliographystyle{IEEEtran}
\bibliography{IEEEexample}

\begin{thebibliography}{10}
\providecommand{\url}[1]{#1}
\csname url@rmstyle\endcsname
\providecommand{\newblock}{\relax}
\providecommand{\bibinfo}[2]{#2}
\providecommand\BIBentrySTDinterwordspacing{\spaceskip=0pt\relax}
\providecommand\BIBentryALTinterwordstretchfactor{4}
\providecommand\BIBentryALTinterwordspacing{\spaceskip=\fontdimen2\font plus
\BIBentryALTinterwordstretchfactor\fontdimen3\font minus
  \fontdimen4\font\relax}
\providecommand\BIBforeignlanguage[2]{{%
\expandafter\ifx\csname l@#1\endcsname\relax
\typeout{** WARNING: IEEEtran.bst: No hyphenation pattern has been}%
\typeout{** loaded for the language `#1'. Using the pattern for}%
\typeout{** the default language instead.}%
\else
\language=\csname l@#1\endcsname
\fi
#2}}

\bibitem{stoffregen1999perceiving}
T.~A. Stoffregen, K.~M. Gorday, Y.-Y. Sheng, and S.~B. Flynn, ``Perceiving
  affordances for another person's actions.'' \emph{Journal of Experimental
  Psychology: Human Perception and Performance}, vol.~25, no.~1, p. 120, 1999.

\bibitem{powers2006advisor}
A.~Powers and S.~Kiesler, ``The advisor robot: tracing people's mental model
  from a robot's physical attributes,'' in \emph{Proceedings of the 1st ACM
  SIGCHI/SIGART conference on Human-robot interaction}, 2006, pp. 218--225.

\bibitem{fussell2008people}
S.~R. Fussell, S.~Kiesler, L.~D. Setlock, and V.~Yew, ``How people
  anthropomorphize robots,'' in \emph{2008 3rd ACM/IEEE International
  Conference on Human-Robot Interaction (HRI)}.\hskip 1em plus 0.5em minus
  0.4em\relax IEEE, 2008, pp. 145--152.

\bibitem{Albrecht2017Reasoning}
S.~V. Albrecht and P.~Stone, ``Reasoning about hypothetical agent behaviours
  and their parameters,'' in \emph{Proceedings of the 16th Conference on
  Autonomous Agents and MultiAgent Systems}, ser. AAMAS '17.\hskip 1em plus
  0.5em minus 0.4em\relax Richland, SC: International Foundation for Autonomous
  Agents and Multiagent Systems, 2017, p. 547–555.

\bibitem{cha2015perceived}
E.~Cha, A.~D. Dragan, and S.~S. Srinivasa, ``Perceived robot capability,'' in
  \emph{2015 24th IEEE International Symposium on Robot and Human Interactive
  Communication (RO-MAN)}.\hskip 1em plus 0.5em minus 0.4em\relax IEEE, 2015,
  pp. 541--548.

\bibitem{jung2013engaging}
M.~F. Jung, J.~J. Lee, N.~DePalma, S.~O. Adalgeirsson, P.~J. Hinds, and
  C.~Breazeal, ``Engaging robots: easing complex human-robot teamwork using
  backchanneling,'' in \emph{Proceedings of the 2013 conference on Computer
  supported cooperative work}, 2013, pp. 1555--1566.

\bibitem{robinson2021smooth}
F.~A. Robinson, M.~Velonaki, and O.~Bown, ``Smooth operator: Tuning robot
  perception through artificial movement sound,'' in \emph{Proceedings of the
  2021 ACM/IEEE International Conference on Human-Robot Interaction}, 2021, pp.
  53--62.

\bibitem{nikolaidis2017game}
S.~Nikolaidis, S.~Nath, A.~D. Procaccia, and S.~Srinivasa, ``Game-theoretic
  modeling of human adaptation in human-robot collaboration,'' in
  \emph{Proceedings of the 2017 ACM/IEEE international conference on
  human-robot interaction}, 2017, pp. 323--331.

\bibitem{chen2018planning}
M.~Chen, S.~Nikolaidis, H.~Soh, D.~Hsu, and S.~Srinivasa, ``Planning with trust
  for human-robot collaboration,'' in \emph{Proceedings of the 2018 ACM/IEEE
  International Conference on Human-Robot Interaction}, 2018, pp. 307--315.

\bibitem{xie2019robot}
Y.~Xie, I.~P. Bodala, D.~C. Ong, D.~Hsu, and H.~Soh, ``Robot capability and
  intention in trust-based decisions across tasks,'' in \emph{2019 14th
  ACM/IEEE International Conference on Human-Robot Interaction (HRI)}.\hskip
  1em plus 0.5em minus 0.4em\relax IEEE, 2019, pp. 39--47.

\bibitem{lee2020getting}
J.~Lee, J.~Fong, B.~C. Kok, and H.~Soh, ``Getting to know one another:
  Calibrating intent, capabilities and trust for human-robot collaboration,''
  \emph{arXiv preprint arXiv:2008.00699}, 2020.

\bibitem{venture2019robot}
G.~Venture and D.~Kuli{\'c}, ``Robot expressive motions: a survey of generation
  and evaluation methods,'' \emph{ACM Transactions on Human-Robot Interaction
  (THRI)}, vol.~8, no.~4, pp. 1--17, 2019.

\bibitem{szafir2014communication}
D.~Szafir, B.~Mutlu, and T.~Fong, ``Communication of intent in assistive free
  flyers,'' in \emph{Proceedings of the 2014 ACM/IEEE international conference
  on Human-robot interaction}, 2014, pp. 358--365.

\bibitem{lemasurier2021methods}
G.~Lemasurier, G.~Bejerano, V.~Albanese, J.~Parrillo, H.~A. Yanco, N.~Amerson,
  R.~Hetrick, and E.~Phillips, ``Methods for expressing robot intent for
  human--robot collaboration in shared workspaces,'' \emph{ACM Transactions on
  Human-Robot Interaction (THRI)}, vol.~10, no.~4, pp. 1--27, 2021.

\bibitem{dragan2013generating}
A.~Dragan and S.~Srinivasa, ``Generating legible motion,'' 2013.

\bibitem{stulp2015facilitating}
F.~Stulp, J.~Grizou, B.~Busch, and M.~Lopes, ``Facilitating intention
  prediction for humans by optimizing robot motions,'' in \emph{2015 IEEE/RSJ
  international conference on intelligent robots and systems (IROS)}.\hskip 1em
  plus 0.5em minus 0.4em\relax IEEE, 2015, pp. 1249--1255.

\bibitem{felis2015optimal}
M.~L. Felis, K.~Mombaur, and A.~Berthoz, ``An optimal control approach to
  reconstruct human gait dynamics from kinematic data,'' in \emph{2015 IEEE-RAS
  15th International Conference on Humanoid Robots (Humanoids)}.\hskip 1em plus
  0.5em minus 0.4em\relax IEEE, 2015, pp. 1044--1051.

\bibitem{liu2005learning}
C.~K. Liu, A.~Hertzmann, and Z.~Popovi{\'c}, ``Learning physics-based motion
  style with nonlinear inverse optimization,'' \emph{ACM Transactions on
  Graphics (TOG)}, vol.~24, no.~3, pp. 1071--1081, 2005.

\bibitem{nicolescu2001learning}
M.~N. Nicolescu and M.~J. Mataric, ``Learning and interacting in human-robot
  domains,'' \emph{IEEE Transactions on Systems, man, and Cybernetics-part A:
  Systems and Humans}, vol.~31, no.~5, pp. 419--430, 2001.

\bibitem{takayama2011expressing}
L.~Takayama, D.~Dooley, and W.~Ju, ``Expressing thought: improving robot
  readability with animation principles,'' in \emph{Proceedings of the 6th
  international conference on Human-robot interaction}, 2011, pp. 69--76.

\bibitem{kwon2018expressing}
M.~Kwon, S.~H. Huang, and A.~D. Dragan, ``Expressing robot incapability,'' in
  \emph{Proceedings of the 2018 ACM/IEEE International Conference on
  Human-Robot Interaction}, 2018, pp. 87--95.

\bibitem{schulman2013finding}
J.~Schulman, J.~Ho, A.~X. Lee, I.~Awwal, H.~Bradlow, and P.~Abbeel, ``Finding
  locally optimal, collision-free trajectories with sequential convex
  optimization.'' in \emph{Robotics: science and systems}, vol.~9, no.~1.\hskip
  1em plus 0.5em minus 0.4em\relax Citeseer, 2013, pp. 1--10.

\bibitem{10.5555/2125842}
R.~Diankov, ``Automated construction of robotic manipulation programs,'' Ph.D.
  dissertation, USA, 2010.

\bibitem{kaelbling2011hierarchical}
L.~P. Kaelbling and T.~Lozano-P{\'e}rez, ``Hierarchical task and motion
  planning in the now,'' in \emph{2011 IEEE International Conference on
  Robotics and Automation}.\hskip 1em plus 0.5em minus 0.4em\relax IEEE, 2011,
  pp. 1470--1477.

\bibitem{morgenstern1953theory}
O.~Morgenstern and J.~Von~Neumann, \emph{Theory of games and economic
  behavior}.\hskip 1em plus 0.5em minus 0.4em\relax Princeton university press,
  1953.

\bibitem{madsen2000measuring}
M.~Madsen and S.~Gregor, ``Measuring human-computer trust,'' in \emph{11th
  australasian conference on information systems}, vol.~53.\hskip 1em plus
  0.5em minus 0.4em\relax Citeseer, 2000, pp. 6--8.

\bibitem{malle2021multidimensional}
B.~F. Malle and D.~Ullman, ``A multidimensional conception and measure of
  human-robot trust,'' in \emph{Trust in Human-Robot Interaction}.\hskip 1em
  plus 0.5em minus 0.4em\relax Elsevier, 2021, pp. 3--25.

\end{thebibliography}

\end{document}